\documentclass[a4paper,10pt,twocolumn,english]{article}

\usepackage{babel}
\usepackage[utf8]{inputenc}
\usepackage[T1]{fontenc}

\usepackage{amsfonts,amssymb,amsmath,amsthm}
\usepackage{subfigure}
\usepackage{graphicx}
\usepackage[footnotesize]{caption}

\usepackage{multirow}
\usepackage{float}
\usepackage{graphicx}
\usepackage{bm}

\usepackage{braket}
\usepackage[ruled,vlined]{algorithm2e}

\usepackage[top=1.5cm, left=1.5cm, right=1.5cm, bottom=1.5cm]{geometry}

\newcommand{\R}{{\mathbb R}}

\newcommand{\be}{\begin{equation}}
\newcommand{\ee}{\end{equation}}
\newcommand{\ba}{\begin{array}}
\newcommand{\ea}{\end{array}}
\newcommand{\baa}{\left[\begin{array}}
\newcommand{\eaa}{\end{array}\right]}

\newcommand{\beqa}{\begin{eqnarray}}
\newcommand{\eeqa}{\end{eqnarray}}
\newcommand{\bt}{\begin{tabular}}
\newcommand{\et}{\end{tabular}}

\newcommand{\bi}{\begin{itemize}}

\newcommand{\ei}{\end{itemize}}
\newcommand{\bc}{\begin{center}}
\newcommand{\ec}{\end{center}}

\newcommand{\norm}[1]{\left\lVert#1\right\rVert}

\newcommand{\eref}[1]{(\ref{#1})}

\title{Unsupervised Dictionary Learning for Anomaly Detection}

\author{
    Paul Irofti$^{1,2}$, Andra Băltoiu$^2$ \\
	\footnotesize $^1$ Computer Science Dept., University of Bucharest, Romania,
    \footnotesize $^2$ The Research Institute of the University of Bucharest (ICUB), Romania}

\date{\empty} 

\renewenvironment{abstract}{\bf\small {\em\ Abstract---}}{}

\begin{document}

\maketitle

\begin{abstract} We investigate the possibilities of employing dictionary learning to address the requirements of most anomaly detection applications, such as absence of supervision, online formulations, low false positive rates. We present new results of our recent semi-supervised online algorithm, TODDLeR, on a anti-money laundering application. We also introduce a novel unsupervised method of using the performance of the learning algorithm as indication of the nature of the samples.
\end{abstract}

\section{Introduction}

It is often the case in anomaly detection applications that not all types of anomalies are known in advance and new ones must be identified on the run.
Supervised methods fall short of this requirement.
We present two approaches on using the dictionary learning (DL) framework for anomaly detection.
The first, a semi-supervised online solution that allows for discovering of new, unseen types of anomalies is presented in Section \ref{sec:semi}.
Previous tests showed the method is suited for malware identification~\cite{IB19_toddler} in large datasets of file descriptions.
The second is an empirical take on the unsupervised problem, created with the application of fraud detection in mind and is detailed in Section \ref{sec:ad}. More on the motivation and specifics of the application, in particular on anti-money laundering, can be found in ~\cite{IPA19_amlsoa} and \cite{BPI19_aml}. 
While the two approaches address slightly different settings, we are motivated in presenting them together by their common novelty in approaching the anomaly detection problem by dictionary learning.

Consider the dictionary learning problem 
\be
\begin{aligned}
& \underset{\bm{D}, \bm{X}}{\min}
& & \norm{\bm{Y}-\bm{DX}}_F^2 \\
& \text{s.t.}
& & \norm{\bm{x}_\ell}_{0} \leq s,\ \ell = 1:N \\
& & & \norm{\bm{d}_j} = 1, \ j = 1:n
\end{aligned}
\label{dict_learn}
\ee
where $\bm{Y} \in \R^{m \times N}$ are the $N$ samples, $\bm{D} \in \R^{m \times n}$ is the dictionary and $\bm{X} \in R^{n \times N}$ the sparse representation, with at most $s$ nonzero elements.

\section{Semi-supervised Anomaly Detection}
\label{sec:semi}

Semi-supervised DL algorithms usually start with an offline pretraining stage where a classification algorithm is used to train a dictionary on labeled data.
This dictionary is then used unsupervised to classify and learn from incoming signals in an online fashion.

Label Consistent K-SVD (LC-KSVD)~\cite{JLD11} extends the DL objective to include a linear classifier $\bm{W}$ and a second label-consistent dictionary $\bm{A}$ that groups atoms to represent only a particular signal class.
\be
\min_{\bm{D},\bm{W}, \bm{A}, \bm{X}} \|\bm{Y}-\bm{DX}\|_{F}^2 + \alpha \|\bm{H}-\bm{WX}\|_{F}^2
+ \beta \|\bm{Q}-\bm{AX}\|_{F}^2
\label{opt_label_cons}
\ee
Here $\bm{H}$ and $\bm{Q}$ are indicator matrices
of the labels and, respectively,
the atoms allocated for representing each signal class.
Due to the properties of the Frobenius norm,
the \eref{opt_label_cons} objective can be rewritten as the plain DL problem.
\be
\min_{\bm{D},\bm{W},\bm{A},\bm{X}}
\left\| \baa{c} \bm{Y} \\ \sqrt{\alpha} \bm{H} \\ \sqrt{\beta} \bm{Q} \eaa -
   \baa{c} \bm{D} \\ \sqrt{\alpha} \bm{W} \\ \sqrt{\beta} \bm{A} \eaa \bm{X} \right \|_F^2
\label{opt_label_cons_ksvd}
\ee

Anomaly detection implies large sets of data where only a few of the items are anomalous.
Thus classic DL solutions can not cope with the large signals set and we have to approach the problem online.
RLS-DLA~\cite{SE10_rls}
is an online recursive algorithm drawing its inspiration from MOD~\cite{EAH99mod}
that fixes the signals and the representations
and updates the dictionary through least squares~(LS).
RLS-DLA applies the same technique in the online setting.
Let us note the following fixed matrices
\be
\bm{G} = \bm{X} \bm{X}^T, \ \ 
\bm{P} = \bm{Y} \bm{X}^T,
\label{GPXY}
\ee
such that the LS at time $t$ is written as
$\bm{D}^{(t)}\bm{G} = \bm{P}$.
When a new signal $\bm{y}$ arrives at time $t+1$,
RLS-DLA performs dictionary update by solving
\be
\bm{D}^{(t+1)} (\bm{G}+ \bm{x} \bm{x}^T ) = \bm{P} + \bm{y} \bm{x}^T.
\label{D_MOD_GP_next}
\ee
After a few algebraic manipulations,
the update can be rewritten as
\be
\bm{D} \leftarrow \bm{D} + \alpha \bm{r} \bm{u}^T
\label{rls_up}
\ee
where 
$\bm{u} = \varphi^{-1} \bm{G}^{-1} \bm{x}$,
$\alpha = \frac{1}{1 + \bm{x}^T \bm{u} }$,
$\bm{r} = \bm{y} - \bm{D} \bm{x}$
and $\varphi$ is a forgetting factor that controls the reliance on past estimations and is usually set around $0.95$.

TODDLeR~\cite{IB19_toddler}
further extends the online objective to include the classification ingredients from LC-KSVD
\be
\begin{aligned}
\min_{\bm{D},\bm{W}, \bm{A}} &\|\bm{y}-\bm{Dx}\|_{2}^2
+ \alpha \|\bm{h}-\bm{Wx}\|_{2}^2 + \beta \|\bm{q}-\bm{Ax}\|_{2}^2 \\
& + \lambda_1\norm{\bm{W}-\bm{W}_0}^2_F + \lambda_2\norm{\bm{A}-\bm{A}_0}^2_F
\end{aligned}
\label{opt_toddler}
\ee
where we also added two regularization constraints on the linear classifier and the label consistent dictionary such that the existing model absorbs the new information given by signal $\bm{y}$ but also maintains its existing properties.
Unlike competing online algorithms~\cite{MB16_lcrls,Zhang13_SemiDiscrim},
TODDLeR updates the dictionary with all samples that it sees no matter the classification confidence level.
The robustness of the existing model is maintained by mediating the rate of change through the two extra regularization in \eref{opt_toddler} that lead to minimizing the following functions after the classification and dictionary update.
Note, however, that since we are in the online, unsupervised stage, labels $h$ are now estimated, not ground truth labels.
\be
f(\bm{W}) = \norm{\bm{h}-\bm{Wx}}_{2}^2 + \lambda_1\norm{\bm{W}-\bm{W}_0}^2_F
\label{opt_W}
\ee
\be
g(\bm{A}) = \norm{\bm{q}-\bm{Ax}}_{2}^2 + \lambda_2\norm{\bm{A}-\bm{A}_0}^2_F
\label{opt_A}
\ee
Looking at $f$ and $g$ as generalized Tikhonov regularization,
we proposed and showed in~\cite{IB19_toddler} that
good choices for $\lambda_1$ and $\lambda_2$
are
\be
\lambda_{1,2}=\norm{\bm{G}}_2 \ \ \text{or} \ \
\lambda_1=\norm{\bm{W}_0}_2, \ \lambda_2=\norm{\bm{A}_0}_2.
\label{lambda_up}
\ee

\section{Unsupervised Anomaly Detection}
\label{sec:ad}
In absence of labels, some other knowledge must be sought that informs DL on the nature of the signals. 
In the following experiments we seek for such clues in the learning process itself.
The scheme involves progressively filtering out signals that, according to some criteria, appear less likely to be anomalies. 
Let $\mathcal{A}$ denote the set of potential anomalies, at first containing all the samples. 
At each iteration the signals not satisfying the criteria are eliminated from the set.

Our first proposal seeks to overspecialize the dictionary in representing the normal samples.  
We train a new dictionary, $\bm{D}_{i}$, on the set $\mathcal{A}$ at each iteration and join these dictionaries together. 
Given that in most applications normal signals outnumber anomalies, it is to be expected that in the first DL iterations the dictionary is more able to represent normal samples. 
Mean representation error obtained after a few rounds of DL can therefore offer some indication on how well the dictionary performs in representing anomalies, and can in turn be used to determine the signal class.
Algorithm \ref{alg:ADDL} details the above steps.

\begin{algorithm}
	\caption{AD-DL}
	\label{alg:ADDL} 	
	\DontPrintSemicolon
	\medskip
	\KwData {samples, $\bm{Y} \in \mathcal{R}^{m \times N}$
		\newline number of global iterations $I$, number of DL iterations $k$}
	\KwResult{label estimates, $l$}

	\medskip
	\nl{Initialize: $\bm{D} = \emptyset $, $\mathcal{A} = \bm{Y}$}, estimates = $\bm{0}_N$\;
	\nl\For {$i = 1:I$}{
		\nl Initialize iteration dictionary $\bm{D}_{i}$ \;
		\nl Update $\bm{D}_{i}$ by performing $k$ iterations of DL on $\mathcal{A}$\;
		\nl Compose $\bm{D} = \bm{D} \cup \bm{D}_{i}$ \;
		\nl Compute representations $X$ using OMP\;
		\nl Compute representation errors $e_i = \Vert \bm{y}_i - \bm{D}\bm{x}_i \Vert_F$ \;
		\nl\If {$i = 1$ } {
			\nl	Compute error threshold  $ e_{mean} = \frac{1}{N} \sum_{i = 1}^{N} e_i$\;		
		}
		\nl Update $\mathcal{A} = \Set{\bm {y}_i \mid e_i > e_{mean}}$  \;
	}
	\nl Label estimates $l_i = 1$ if $\bm {y}_i  \in \mathcal{A}$
\end{algorithm}

The dictionary can, however, overspecialize in representing some anomalies as well, which will then be excluded from $\mathcal{A}$.
Therefore, a criteria that minimizes the number of false negatives is also needed. 
We turn to the measure of atom popularity for the task. 
The approach requires that the number of anomalies in the dataset, $N_a$, is known, which is reasonably common in practice. 
Provided the dictionary has learned to represent both the features of normal signals and those of anomalies, it is expected that some dictionary atoms will specialize in describing anomalies. 
As a result, those atoms will be used in the representations of at most $N_a$ samples.
We refer to popularity of an atom $j$ as the number of signals that are represented using that atom and compute the measure as
$p_j = \Vert \bm{x}_j^\top \Vert_0$ \cite{dl_book}.

In our second solution, we restrict the set $\mathcal{A}$ of potential anomalies to signals represented by atoms $\bm{d}_j$ with popularity $p_j > N_a$, namely $\mathcal{A}= \Set{\bm{y}_k \mid x_{j,k} \ne 0 \wedge p_j > N_a, \forall j,k}$. Atoms that describe rare but otherwise normal features will also meet the criteria, therefore filtering out signals using the above atom popularity threshold is a cautious way of reducing the set of potential anomalies: it minimizes false negatives, but may result in fairly large number of false positives.
Unlike the previous method, a new dictionary is learned at each step.
The iterative process can be stopped when $\vert \mathcal{A} \vert = N_a$.

\section{Results}
We test the above methods on a financial fraud database consisting of credit card transaction information developed at the Université Libre de Bruxelles \cite{Car19_kccf}. The database is extremely unbalanced ($0.17 \%$ anomalies) and for the purpose of this study we work on a subset where the normal samples outnumber anomalies by a factor of 100 when testing the online method (TODDLeR) and 10 for the unsupervised methods.

We first run TODDLeR on a dataset of $N = 39754$ samples with 29 features, which have been previously normalized. The best performance, $98.86 \%$ classification accuracy, is obtained when the Tikhonov regularization factors are set  $\lambda_{1,2}=\norm{\bm{G}}_2$. 
This setting also has a good false positive count of 83.

\begin{figure}
\centering
\includegraphics[width=0.7\linewidth]{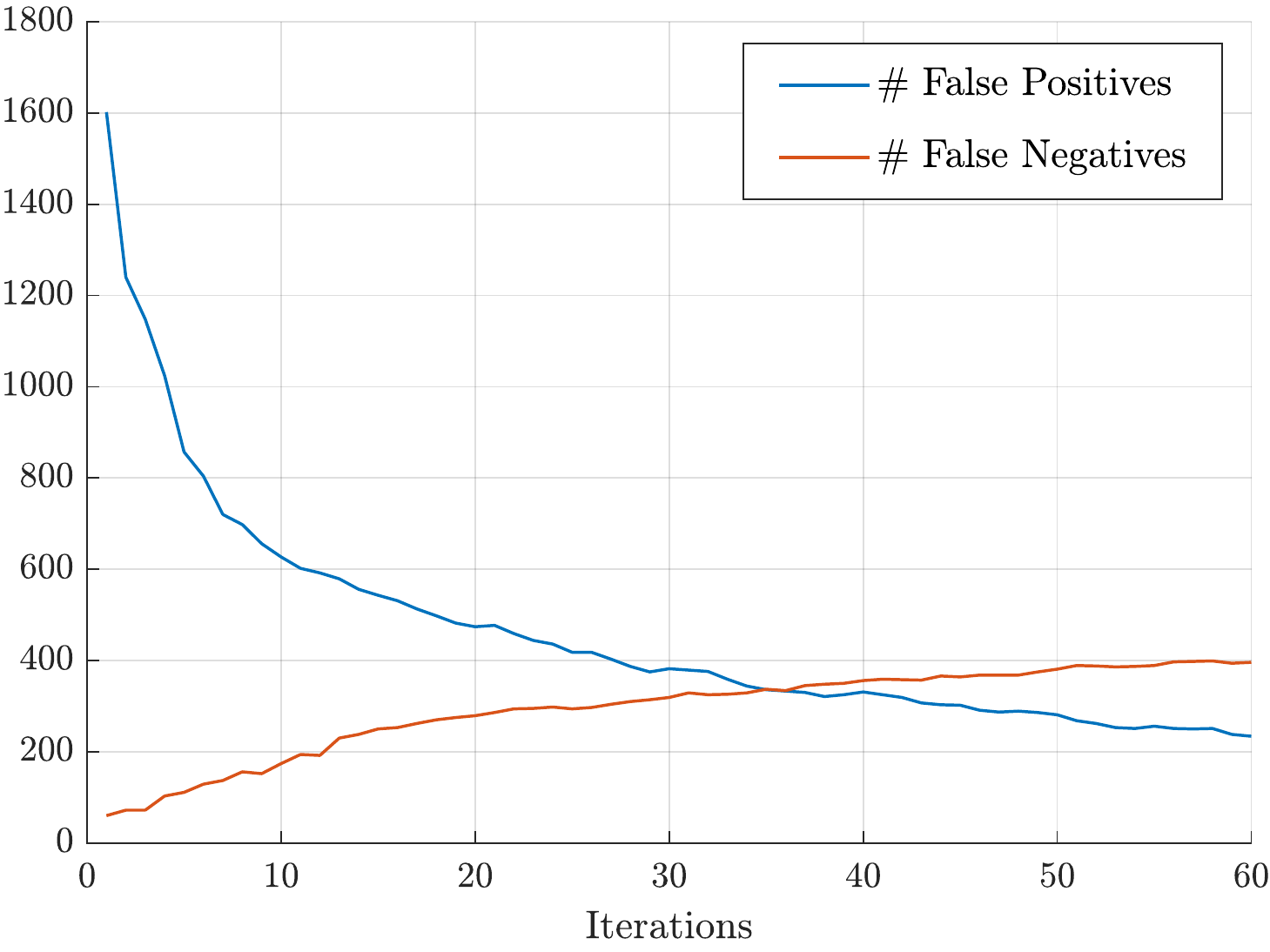}
\caption{Unsupervised DL with error threshold}
\label{fig:addl_mean}
\end{figure}

Our second test involves unsupervised experiments on a dataset of $N = 5412$ samples.
At each step, we perform 20 iterations of DL (AK-SVD). 
Sparse coding is done via OMP \cite{PRK93omp} with sparsity set to $s = 5$.
Figure \ref{fig:addl_mean} shows the evolution of false positives and false negatives when filtering samples based on the error criteria. 
Filtering with popularity threshold alone correctly labels $32\%$ of signals as normal samples, while keeping false negatives to 0 after 60 iterations. Clearly, performance depends on the number of atoms representing other uncommon signal attributes. 
A preliminary step of dictionary size adjustment can help ensure that a proper number of atoms is available to represent all sample features.
This empirical approach can be extended to other means of characterizing the learning process (such as atom coefficients magnitude, or the evolution of these measures as learning progresses etc) that can be informative of the nature of the signals or their occurrence rate.

\section{Conclusion}
\label{sec:conc}

The paper shows current work state on the problem of applying DL to anomaly detection.
It tackles the main issues arising in these applications, namely the need for lightweight, online algorithms, with little or no supervision. 

\section*{Acknowledgements}

This work was supported by BRD Groupe Societe Generale through DataScience Research Fellowships of 2019.
P. Irofti was  also supported  by  a  grant  of  Romanian  Ministry  of  Research  and  Innovation  CCCDI-UEFISCDI.  project  no.  17PCCDI/2018,
and
A. Băltoiu by the Operational Programme Human Capital of the
Ministry of European Funds through the Financial Agreement
51675/09.07.2019, SMIS code 125125.
\bibliography{itwist20}
\bibliographystyle{plain}

\end{document}